\documentclass[conference]{IEEEtran}
\ifCLASSINFOpdf
\else
\fi

\usepackage{algorithm}
\usepackage{algorithmic }
\usepackage{float}
\usepackage{amssymb}
\usepackage{hhline}
\usepackage[table]{xcolor}

\usepackage{array}
\usepackage{multirow}
\usepackage{url}
\usepackage[table]{xcolor}

\usepackage{tablefootnote}

\usepackage{enumitem}

\usepackage{pdfpages}

\usepackage{tikz}

\usepackage{xfrac}
\usepackage{fixltx2e}

\begin{document}
%
\title{Skilled Experience Catalogue: A Skill-Balancing Mechanism for Non-Player Characters using Reinforcement Learning}

\author{\IEEEauthorblockN{Frank G. Glavin}
\IEEEauthorblockA{School of Computer Science, \\
National University of Ireland, Galway. \\
Email: frank.glavin@nuigalway.ie}
\and
\IEEEauthorblockN{Michael G. Madden}
\IEEEauthorblockA{School of Computer Science, \\
National University of Ireland, Galway. \\
Email: michael.madden@nuigalway.ie}
}


%


\maketitle

\begin{abstract}
In this paper, we introduce a skill-balancing mechanism for adversarial non-player characters (NPCs), called \emph{Skilled Experience Catalogue} (SEC). The objective of this mechanism is to approximately match the skill level of an NPC to an opponent in real-time. We test the technique in the context of a First-Person Shooter (FPS) game. Specifically, the technique adjusts a reinforcement learning NPC's proficiency with a weapon based on its current performance against an opponent. Firstly, a catalogue of experience, in the form of stored learning policies, is built up by playing a series of training games. Once the NPC has been sufficiently trained, the catalogue acts as a timeline of experience with incremental knowledge milestones in the form of stored learning policies. If the NPC is performing poorly, it can jump to a later stage in the learning timeline to be equipped with more informed decision-making. Likewise, if it is performing significantly better than the opponent, it will jump to an earlier stage. The NPC continues to learn in real-time using reinforcement learning but its policy is adjusted, as required, by loading the most suitable milestones for the current circumstances.
\end{abstract}
\section{Introduction}
\indent This paper presents a new mechanism for Dynamic Difficulty Adjustment in the context of reinforcement learning called Skilled Experience Catalogue (SEC). Specifically, we store the current policy of the non-player character (NPC) at various intervals during an initial training phase. Once the training phase is complete, the base policy of the NPC can be adjusted in real-time, influenced by a threshold value, to approximately match the skill level of the current opponent. \\
\indent To test our SEC mechanism, we apply it to the weapon proficiency of an NPC bot in a First-Person Shooter (FPS) \emph{Deathmatch} game. Specifically, the mechanism applies only to the learned task of aiming/firing a weapon at an enemy. The NPC has fixed strategies for the other in-game tasks such as item collection, opponent evasion and travelling around the map. The NPC bot is initially trained against a single opponent and builds up a catalogue of reinforcement learning policies as it gains experience from using the weapon over time. These stored policies, that loosely represent skill level, can then be loaded in subsequent games to balance the gameplay against the current opponent. We demonstrate this SEC mechanism against five different levels of fixed-strategy opponents and show that a single catalogue of experience can be used to closely match the performance of each. The NPC that we have developed is an adversarial one \cite{treanor2015ai} which contrasts with supportive companion NPCs \cite{guckelsberger2016intrinsically} found in some game genres.\\
\indent The approach that we present is novel in that it is using a by-product of the bot's learning process to create \emph{milestones} which represent the knowledge acquired at the different stages of learning. The policies of the bot are stored, at different stages, to keep a sequential catalogue of experience. The bot can then jump to the most appropriate policy to coincide with the skill level of the current opponent while continuing to adapt based on its in-game experience. Our approach does not require manually optimising parameters to represent different skill levels. Conversely, we are sampling from the natural learning progress of the agent over time.
\section{Background Information}
\subsection{Dynamic Difficulty Adjustment}
Traditionally, human computer game players select a difficulty setting from a menu before beginning the game. This can be as simple as selecting \emph{easy} / \emph{medium} / \emph{hard} or can include more detailed options for the player to choose from. These settings have a direct, and usually static, effect on the skill level of the NPCs. Such fixed-difficulty settings can often be too broad. For instance, a setting that is intended to be easy may nonetheless be too difficult for some players. Players may also improve their performance at different rates. The traditional approach does not make use of the player's current performance measures to direct the gameplay. While this approach has the benefit of simplicity, from a game development point of view, it can lead to predictable gameplay when static rule-based opponents are deployed which can adversely affect the entertainment value of the game. Dynamic Difficulty Adjustment (DDA) \cite{hunicke2004ai}, which can also be referred to as Dynamic Game Balancing (DGB) \cite{DGB}, involves identifying the player's performance and skill level, and then dynamically adjusting the difficulty level accordingly. The goal of this is to ensure that the game remains challenging and can cater for many different players of varying skill levels.
\subsection{Reinforcement Learning}
Reinforcement learning (RL) is a branch of artificial intelligence in which a learner, often called an \emph{agent}, interacts with an environment to achieve an explicit goal or goals \cite{sutbar}. The environment consists of a  set of states, called the \emph{state space}, and the agent must choose an available action from the \emph{action space} when in a given state at each time step. The agent learns from its interactions with the environment, receiving feedback for its actions in the form of numerical rewards, and aims to maximise the reward values that it receives over time. Two common approaches to storing/representing a \emph{policy} in reinforcement learning are \emph{generalisation} and \emph{tabular}. With generalisation, a function approximator is used to generalise a mapping of states to actions. The tabular approach, which is used in our research, stores numerical representations of all state-action pairs in a lookup table. The agent's decision-making involves choosing between exploring the effects of taking novel actions and exploiting the knowledge that it has acquired from earlier exploration. Reinforcement learning is inspired by the process by which humans interact with the world and learn from experience.
\subsection{Game Environment and Development Tools}
\indent For this research, we use the game Unreal Tournament 2004 (UT2004) which is a commercial FPS game \cite{ut2004}. The agents are developed using a toolkit called \emph{Pogamut 3} \cite{pogamut} which is an open-source development platform for creating virtual agents in the 3D game environment of UT2004. The main objective of Pogamut 3 is to simplify the coding of actions taken in the environment, such as path finding, by providing a modular development platform. Making use of these primitives, the focus of our development is on producing intelligent NPC behaviour.
\section{Motivation}
When computer-controlled opposition is too strong, human players can become frustrated with the gameplay. Conversely, opponents that are too weak result in predictable games in which human players do not feel challenged \cite{koster2013theory}. The challenge of a game is widely considered to play a crucial role in the player's overall enjoyment \cite{challenge}. We believe that successful game AI requires techniques to be developed in which the NPCs can learn good tactics independently as well as being both unpredictable and adaptive to their surroundings. Keeping a player's win and loss rate close and unpredictable in a game can increase the player's overall suspense and the game's outcome uncertainty. Abuhamdeh \emph{et al.} \cite{abuhamdeh2015enjoying} carried out a study on the relevance of outcome uncertainty and suspense for intrinsic motivation and concluded that games with higher outcome uncertainty were more enjoyable to play.\\
\indent We observe that modern computer games can often lack flexibility with regards to difficulty settings and this can lead to mismatches between the player's ability and the overall difficulty of the game. DDA can be used to ease the learning process for beginners. Difficulty settings are balanced in real-time in contrast to traditional approaches that involve extensive user testing and redesign in order to identify suitable levels. This can be a costly and time-consuming process \cite{rollings}. \\

\section{Related Research}
\indent Hunicke and Chapman \cite{hunicke2004ai} presented an interactive DDA system called \emph{Hamlet} which is an integrated set of libraries within the \emph{Half-Life} SDK. The Hamlet system has an \emph{evaluation function}, which maps the current state of the game world to an evaluation of the player's performance and an \emph{adjustment policy}, for mapping the evaluation to adjustments in the game world. Hamlet monitors incoming game data and estimates the player's future state from the data. If an undesirable state is predicted, the system will intervene and adjusts the game settings as required. Hunicke \cite{hunicke05} used Hamlet to examine the requirements for incorporating effective dynamic difficulty adjustment into an FPS game. The aim of the study was to identify if DDA could be performed effectively, without degrading the core gameplay experience for the user. The authors reported that their preliminary results show an improvement in player performance, while retaining the player's sense of agency and accomplishment. \\
\indent Spronck et al. \cite{spronck} showed the extent to which their technique of \emph{dynamic scripting} \cite{spronck-2} could be used to adapt game AI to balance the gameplay in a simulation closely related to the Role-Playing Game (RPG) \emph{Baldur's Gate}. The authors focussed on enhancing the difficulty-scaling properties of the dynamic scripting technique. These were high-fitness penalising, weight clipping, and top culling. The \emph{reward peak value} in dynamic scripting determines how effective the opponent behaviour will be. With high-fitness penalising, this value is adjusted after every fight depending on the outcome. If the computer-controlled opponent wins, it is reduced; otherwise it is increased. There is also a maximum and minimum value that this reward peak value can be. The \emph{maximum weight value} determines the maximum level of optimisation a learned tactic can achieve. With weight clipping, this value is automatically changed to balance the overall gameplay. Top culling is similar to weight clipping, however, rules with a weight greater than the maximum weight value are allowed. Those that exceed the maximum weight value will not be selected for a generated script which will force the computer-controlled opponent to use weaker tactics. The authors reported that, of the three different difficulty-scaling enhancements, the top-culling enhancement was the best choice. It was reported that it produced results with low variance, was easily implemented, and was the only one of the three enhancements that managed to force a balanced game when inferior tactics were used.\\
\indent Tan et al. \cite{tan} presented two adaptive algorithms, based on ideas from reinforcement learning and evolutionary computation, to scale the difficulty of the game AI to improve player satisfaction. They introduced two controllers, namely, the adaptive uni-chromosome controller (AUC) and the adaptive duo-chromosome controller (ADC). The authors examined the effects of varying the learning and mutation rates and proposed general rules for setting these parameters. The authors carried out experimentation using a modified version of the car simulator used in the Simulated Car Racing Competition held during the 2007 IEEE Congress on Evolutionary Computation (CEC 2007). It was demonstrated that the proposed algorithms can match their opponents in terms of mean scores and winning percentages. The authors also reported that both algorithms were able to generalize well to a variety of opponents. \\
\indent Vicencio-Moriera et al. \cite{aimAssist} carried out three separate studies on the performance of different \emph{aim assist} techniques in an FPS game developed by the authors using the Unreal Development Kit (UDK). Aim assistance in FPS games is used to make it easier for players to select on-screen targets. This enables players with less skill to perform at a more competitive level increasing the competition and, in turn, the enjoyment of the player, which essentially balances the gameplay. The authors compared the following aim assist techniques: \emph{target lock} (moves the crosshairs of the player's weapon to the closest target's head), \emph{bullet magnetism} (bullets towards the closest target if within the activation range), \emph{area cursor} (physical size of the crosshair changes), \emph{sticky targets} (changes control-to-display ratio when the crosshairs are over a target) and \emph{gravity} (targets have an attractive force dragging crosshairs towards them). The authors reported that the assists worked well in a target-range scenario, but their performance was reduced when real-game elements were introduced. They reported that bullet magnetism and area cursor worked well in a wide variety of situations but techniques such as target lock, while working well, were too perceptible to be successful for balancing. Vicencio-Moriera extended this work, in \cite{FPSbalance}, by refining the bullet magnetism and area cursor techniques. They also developed a new technique to maintain the effect of the aim assist for a longer duration. They also included a map for novice players that shows them the location of expert players and incorporated different rates of damage for stronger and weaker players. The authors report that the new balancing schemes were extremely effective with the ability to balance the gameplay amongst players with large skill differences. \\
\indent Our approach, which is outlined in the following section, differs from the aforementioned methods in that it uses real-time learning during the game coupled with a mechanism for loading policies that closely match the current opponent.
\section{Skilled Experience Catalogue}
\begin{figure*}[h!]
  \begin{center}  
      \includegraphics[width=5in]{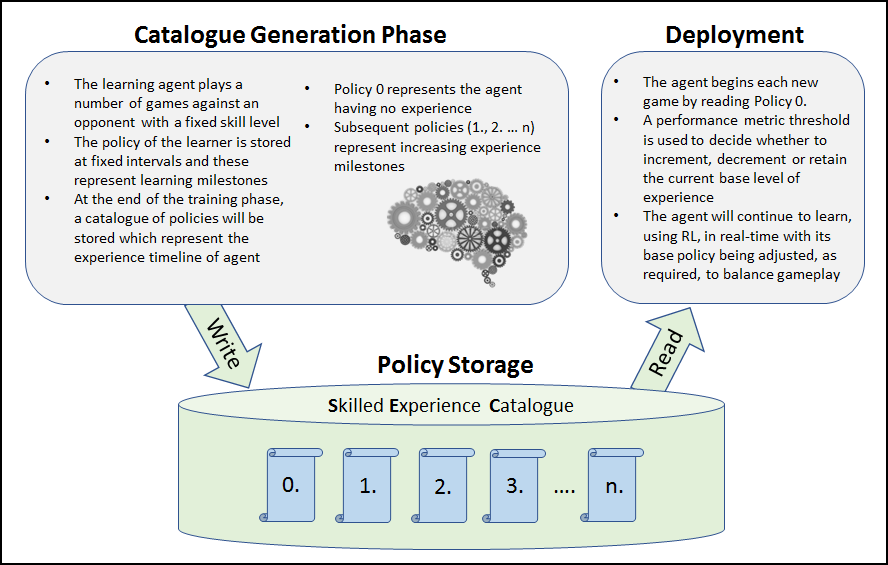}
    \caption{An overview of Skilled Experience Catalogue.}
    \label{secOverview}
  \end{center}
\end{figure*}

We designed SEC based on the premise that there is a progressive timeline which begins with poor performances and ends with good performances as the agent learns how to perform a task over time. In general, SEC involves storing the policy of an RL agent at periodic intervals during an initial training phase. The agent will begin the training phase with no knowledge of the environment or intuition on what the best actions are to take given the circumstances. Well-designed learning agents will depict a clear upward trend in performance over time as the agent gains more experience. This time-based increase in performance is crucial to the success of using SEC. Milestones of the timeline of experience are stored by recording the RL policy at set intervals during the training phase. Once the experience catalogue has been populated with these learning milestones they can then be loaded to either increase or decrease the current underlying ability of the agent by effectively adding or removing experience (see Figure \ref{secOverview}). Our application of SEC is concerned with balancing the \emph{Assault Rifle} skill of a Deathmatch NPC playing against a single opponent. The NPC is initially trained using the shooter bot implementation from Glavin and Madden \cite{glavin-1}\cite{glavin-2}. \\
\begin{figure}[h!]
  \begin{center}  
      \includegraphics[width=3.45in]{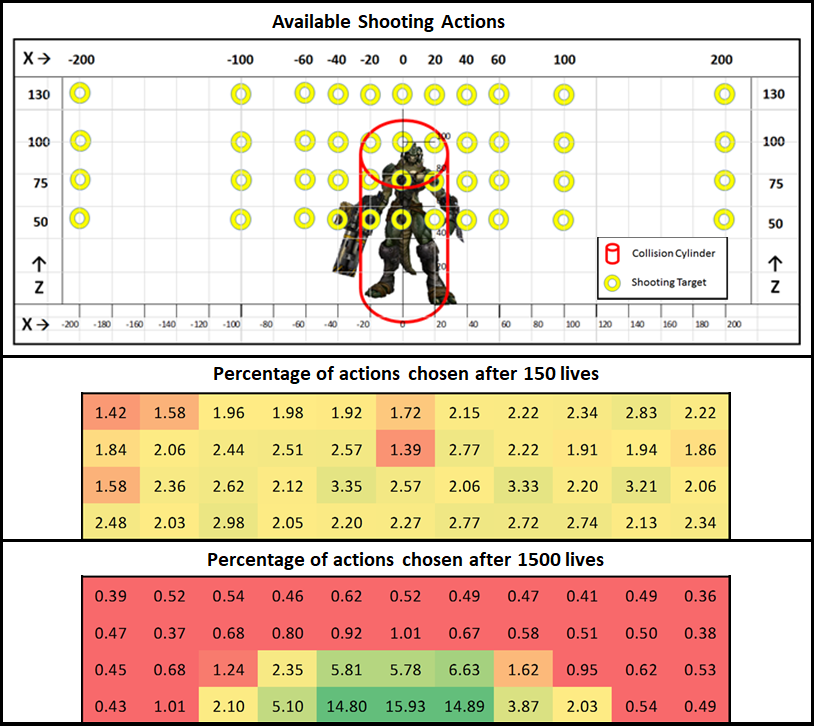}
    \caption{Shooting actions available (top) and varied decision-making based on experience \cite{glavin-2} (middle/bottom). The heat maps represent the percentage of time that a specific action was chosen up to that point. This is illustrated for those chosen up to 150 lives and up to 1500 lives respectively to show the improved decision making.}
    \label{secShooting}
  \end{center}
\end{figure}
\indent The main features of this implementation are as follows. The state space is made up of a series of discretised features including the relative speed, relative moving direction, relative rotation, and distance to the opponent. The velocity and direction values of the opponent are translated into the NPC's point of view so that the NPC's actions can include discretized variations of shooting in a forward direction. The opponent's direction and speed are recorded relative to the NPC’s own speed and from the perspective of the NPC looking directly ahead. Full details on the state space can be found in Glavin and Madden \cite{glavin-2}. \\
\indent Figure \ref{secShooting} shows the actions that are available to the NPC, the percentage of those chosen after 150 lives and the percentage of those chosen after 1500 lives. These values were chosen as representative early and late stage learning to highlight the differences in decision-making as the NPC gains experience. The actions are expressed as different target directions in which the NPC can shoot, and which are skewed from the opponent's absolute location on the map. The amount of skew along the X-axis (left and right) and Z-axis (up and down) varies by different fixed amounts as shown in the upper segment of Figure \ref{secShooting}. These actions were designed specifically for the Assault Rifle weapon. The logic of this shooter NPC is based on the SARSA($\lambda$) \cite{sutbar}\cite{rumm} algorithm which uses eligibility traces \cite{klopf} to speed up learning by allowing past actions to benefit from the current reward. The NPC receives a reward of 250 every time the system records that it has caused damage to the opponent with the shooting action. If the NPC shoots the weapon and does not hit the opponent, it receives of penalty of -1. We use a tabular approach to represent the policy of the learner by storing numerical representations (Q-values) of all state-action pairs in a lookup table (Q-table). These are periodically stored and used to represent base learning levels (policies) as milestones. Full details of this algorithm and NPC implementation can be found in Glavin \cite{glavin-3}. The following sections include details on the training experimentation and milestone switching mechanism for SEC.
\subsection{Training Experience}
The NPC must play a series of games against an opponent to populate the catalogue of experience. Each time the NPC dies it will write out its Q-table to a file. Once this training period has been completed, milestone Q-tables (from set intervals of the learning timeline) must be selected and placed into the SEC. We have chosen 100 deaths to represent a milestone in our implementation. That is to say that a milestone Q-table is added to the SEC after 100 deaths, 200 deaths, 300 deaths and so on. Since continuous reinforcement learning is being used, we believe that the resulting performance will be more natural and will make the adjustments of the skill level harder to detect.
\subsection{Skill Adjustments}
\indent Once the SEC has been populated with the desired number of milestones, the mechanism for changing milestones must be set. Our implementation uses a simple positive/negative threshold for kill-death differences (KDDs) to drive the switching between milestones. Specifically, if the KDD between the NPC and the opponent exceeds a value of 5 (meaning the bot is performing significantly better), the policy will revert back to the previous milestone from the SEC. It will continue to step back through the milestones, after every opponent death, while the KDD remains above 5 or until it has reached the first milestone (empty Q-table; no knowledge). If the KDD falls below -5, the next milestone will be chosen from the SEC after every death until either the KDD returns to the \emph{match range} ($-5 \geqslant KDD \leqslant 5$) or the highest milestone has been reached. The current milestone will remain unchanged while the KDD value is within the match range. The value 5 was chosen after a series of preliminary runs. This value could be increased to make the skill adjustments less prominent or reduced to enable faster skill adjustments. \\
\indent The learning algorithm that is controlling the bot is as described in Glavin and Madden \cite{glavin-2}. Both Periodic Cluster Weighted Rewarding (PCWR) and Persistent Action Selection (PAS) are both enabled, and all of the algorithm settings are as described in that research work.
\section{Experimentation}
\label{secExper}
In this section, we outline how we first trained the NPC by playing against a single opponent for 100 individual thirty-minute games. The purpose of this training phase is to populate the catalogue of experience as the NPC acquires knowledge through trial and error over time. After this, we discuss the experiments in which the NPC used the SEC mechanism, while playing against opponents with differing skill levels, to balance the game play. We then discuss re-running these experiments with the technique disabled to carry out a comparative analysis.
\subsection{Training Experiments}
There are eight different pre-programmed native bot skill levels in UT2004 that are designed to increase the challenge for human players as the skill level is increased. High-level descriptions of the attributes associated with the first five of these skill levels, as reported in \cite{unrealWiki}, are listed in Table \ref{nativeBotTable}.
\begin{table}[h]
\begin{center}
\setlength{\extrarowheight}{0.15cm}
\begin{tabular}{|c|p{6cm}|}
\hline
\bf{Skill} & \bf{Attributes}\\
\hline
\emph{Novice} & 60\% of regular running speed, will not move during combat unless very weak, limited perception with 30$^{\circ}$ field of view, shooting aim can range 30$^{\circ}$ off target, slow to turn. \\
\hline
\emph{Average} & 70\% of regular running speed, slightly higher shooting accuracy, turns slightly faster than novice. \\
\hline
\emph{Experienced} & 80\% of regular running speed, will move and fire simultaneously, 40$^{\circ}$ field of view, can turn by more than $\sfrac{1}{2}$ per second. \\
\hline
\emph{Skilled} & 90\% of regular running speed, can double jump, 60$^{\circ}$ field of view, turns more than $\sfrac{5}{8}$ per second.\\
\hline
\emph{Adept} & Run at full speed, will dodge enemy fire, will close in on enemy, aim ``leads'' the target, 80$^{\circ}$ field of view, turn almost $\sfrac{3}{4}$ per second. \\
\hline
 \end{tabular}
\end{center}
\caption{UT2004 native bot skill attributes. (Source: \cite{unrealWiki})}
\label{nativeBotTable}
\end{table}
The first step of our experimentation involved training the NPC by playing it against a Level 5 (Adept) opponent. We ran one hundred 1-vs-1 Deathmatch (30 minute) games on the Training Day map. This is a small map which encourages almost constant combat. The resulting data showed that the learner NPC performed poorly during the early games in which it died much more often than killing the opponent. 
\begin{figure}[h!]
  \begin{center}  
     \includegraphics[width=3.45in]{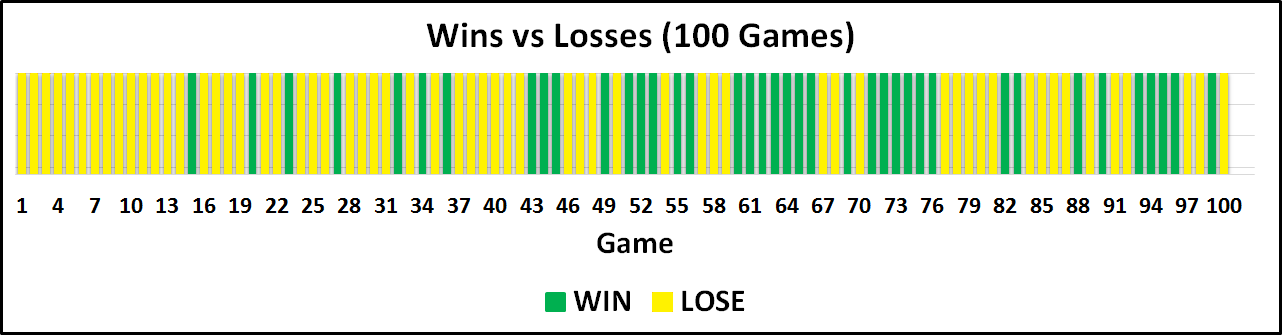}
    \caption{Games won and lost during the one hundred 30 minute training games. The number of games won increases in the latter stages of the training phase.}
    \label{winLose}
  \end{center}
\end{figure}
This would be expected as the NPC needs time to experience all the game states and experiment with the different actions. The NPC began to outperform the opponent at the half way point of the 100 games.
Figure \ref{winLose} shows the number of wins and losses that the NPC recorded against the Level 5 opponent during the 100 training games. The NPC must play 15 games before it manages to defeat the Level 5 opponent. The NPC manages to beat the opponent much more frequently when it passes the half way point of the 100 games. This verifies that the NPC is in fact learning how to play more proficiently as it gathers experience. The NPC wins 39 out of the 100 games with 28 of these wins occurring in the second half of the games played. During the 100 games, the NPC died almost 9000 times. As a result of this we stored 90 milestone Q-tables. These were from no experience (empty Q-table) up to having died 8900 times in increments of 100. \\
\indent Table \ref{Q1vQ4} compares the average kills, deaths and kill-death (KD) ratio from the first 25 games (First Quarter - Q1) to the last 25 games (Fourth Quarter - Q4) in order to test if the enabled learning is leading to a statistically significant improvement in the average performance in the latter stages of learning. The ** in the table entries signifies that there are statistical differences with significance level $\alpha$ equal to 0.05 using an unpaired two-tailed t-test. It is important to note that in each successive game the bot begins with the knowledge that it has built up from all of the previous games. Therefore, the examples are not strictly independent as memory from each game is persisting over time. The settings of the individual 30-minute games and the opponents remain consistent throughout the 100 games. Both sample sets comprise individual games that take place in a period of 12.5 hours of learning.\\
\indent The purpose of this comparison is to check for a statistically significant difference in the average performance between these two learning periods (early and later learning) in order to verify that the bot continues to improve its performance over time. For instance, if there was no significant difference between the bot's average performance during the period of 0 to 12.5 hours and the bot's average performance during the period of 37.5 to 50 hours then we could conclude that the effect of learning on the bot's performance had already plateaued during the earlier stages of learning. Table \ref{Q1vQ4} shows that this is not the case and that the average number of kills achieved, and the average kill-death ratio has improved, at a 95\% confidence level, in the later learning period. We have also confirmed from the bot logs that the bot continues to encounter new states throughout all of the training games and therefore may have an opportunity to learn a better policy in the latter stages of the training phase.
\begin{table}[h!]
\caption{Comparison of performance between Q1 and Q4. Level of confidence: ** = 95\%. This shows an increase in performance in the latter stages of learning.}
\label{Q1vQ4}
\begin{center}
\setlength{\extrarowheight}{0.15cm}
\begin{tabular}{|c|c|c|}
   \hline
& \vtop{\hbox{\strut \bf{Q1: Games 1 to 25}}\hbox{\strut \bf{\hspace{3mm}Avg (Std Error)}}} &\vtop{\hbox{\strut \bf{Q4: Games 75 to 100}}\hbox{\strut \bf{\hspace{3.5mm}Avg (Std Error)}}} \\
\hline
Kills &79.20	($\pm$ 2.09)	&\bf{84.56	($\pm$ 1.46) **} \\
\hline
Deaths &91.60	($\pm$ 2.02)	&87.52	($\pm$ 1.46) \\
\hline
KD Ratio &0.88	($\pm$ 0.04)	&\bf{0.98	($\pm$ 0.03) **} \\
\hline
\end{tabular}
\end{center}
\end{table}
\subsection{SEC-Bot versus Fixed Strategy Opponents}
From here on, we will refer to the NPC that uses the SEC as \emph{SEC-Bot}. The SEC-Bot begins each new game that it plays with no experience and then increases or decreases its knowledge as required, based on the threshold, by using the Q-table milestones. These experiments were also carried out on the Training Day map.\\
\indent Table \ref{winsAndLosses} shows the number of times that the SEC-Bot won, lost and drew against the first 5 levels of the native fixed-strategy bots. The SEC-Bot won 39, lost 30 and drew 6 of the 75 games that were played (15 individual games per level of native opponent) which shows that there is a large amount of outcome uncertainty when playing against five different levels of opponent. For instance, the SEC-Bot does not constantly beat the Level 1 opponent all the time nor does it struggle to play at the same standard as the Level 5 opponent.
\begin{table}[h!]
\caption{Wins, losses and draws for the SEC-Bot against each of the 5 levels.}
\label{winsAndLosses}
\begin{center}
\setlength{\extrarowheight}{0.15cm}
\begin{tabular}{|c|c|c|c|}
   \hline
\bf{Opponent Skill Level}&\bf{Win} &\bf{Lose} &\bf{Draw}  \\
\hline
 Level 1 & 8 & 6 & 1 \\
\hline
 Level 2 & 9 & 6 & 0 \\
\hline
 Level 3 & 8 & 4 & 3 \\
\hline
 Level 4 & 7 & 6 & 2\\
\hline
 Level 5 & 7 & 8 & 0\\
\hline
\end{tabular}
\end{center}
\end{table}
We can see from the table that the win and loss rate for the SEC-Bot are closely matched and that it is managing to successfully balance the gameplay against the different levels of opponent using just a single catalogue of experience. As we will see later, the difference between a win and a loss is often only a small number of kills. The following figures, from Figure \ref{secLevel1} to Figure \ref{secLevel5}, show the results of the SEC-Bot playing a total of 15 thirty-minute games against five different levels of opponent. The purpose of this is to observe how well it can match the opponent's skill level, with respect to killing and being killed, by using the SEC mechanism. From the results, we can see that the SEC-Bot manages to closely match the kill rate of the opponent over each of the different levels. 
\begin{figure}[h!]
  \begin{center}  
      \includegraphics[width=3.45in]{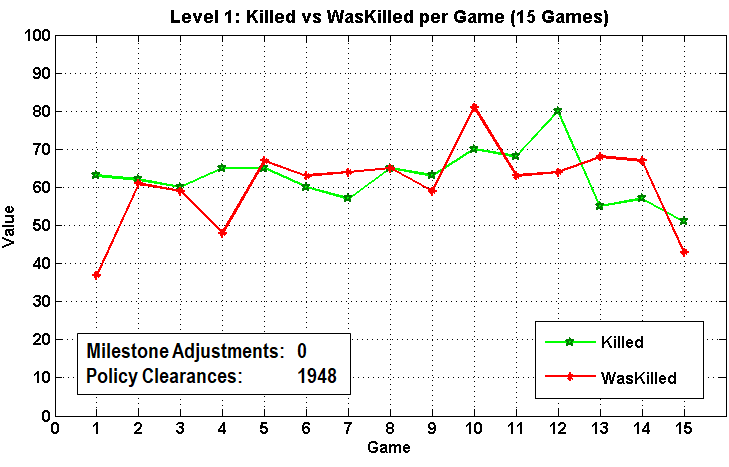}
    \caption{SEC-Bot: Killed vs WasKilled against a Level 1 opponent.}
    \label{secLevel1}
  \end{center}
\end{figure}
During these experiments, we noted the changes to the milestones that were occurring during the game-play. These include any time the bot moved up or down a milestone from the SEC and are listed as \emph{milestone adjustments} on the figures. A \emph{policy clearance} occurs when the SEC-Bot is currently using Policy 0 and still out-performing the opponent. It involves clearing all of the Q-values that were built up during gameplay and essentially re-starting learning. For Level 1, the SEC-Bot never increased to a higher milestone than Policy 0. At the beginning of each game, it started with no knowledge and then only required the in-game learning experience to remain balanced with this opponent. It did, however, have to carry out a policy clearance over 1900 times throughout the games as it was out-performing the Level 1 opponent at the early stages of learning. 
\begin{figure}[h!]
  \begin{center}  
      \includegraphics[width=3.45in]{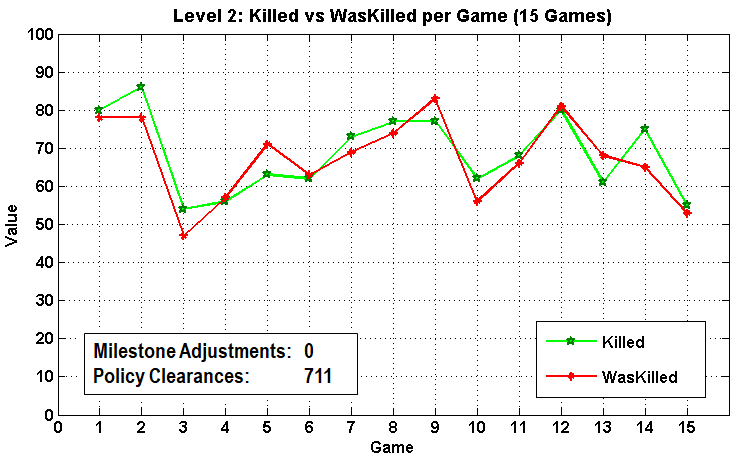}
    \caption{SEC-Bot: Killed vs WasKilled against a Level 2 opponent.}
    \label{secLevel2}
  \end{center}
\end{figure}
The SEC-Bot, once again, had no milestone increases during the Level 2 games whereas the changes were much more prevalent against the more difficult opponents (Level 3, Level 4, and Level 5).
\begin{figure}[h!]
  \begin{center}  
      \includegraphics[width=3.45in]{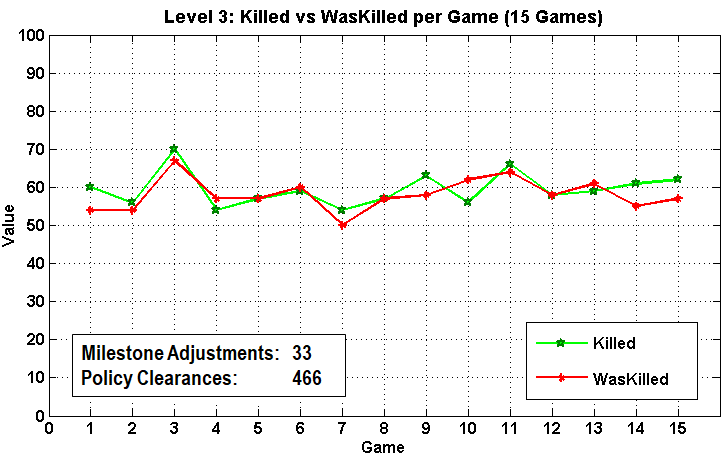}
    \caption{SEC-Bot: Killed vs WasKilled against a Level 3 opponent.}
    \label{secLevel3}
  \end{center}
\end{figure}
\begin{figure}[h!]
  \begin{center}  
      \includegraphics[width=3.45in]{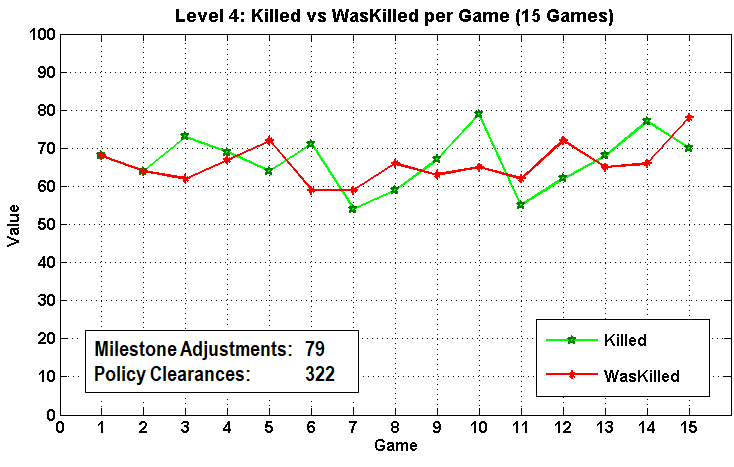}
    \caption{SEC-Bot: Killed vs WasKilled against a Level 4 opponent.}
    \label{secLevel4}
  \end{center}
\end{figure}
The SEC-Bot had to almost immediately rise up through the milestones (once it fell below the threshold), before stopping at the highest point, while playing against the Level 5 opponent.
\begin{figure}[h!]
  \begin{center}  
      \includegraphics[width=3.45in]{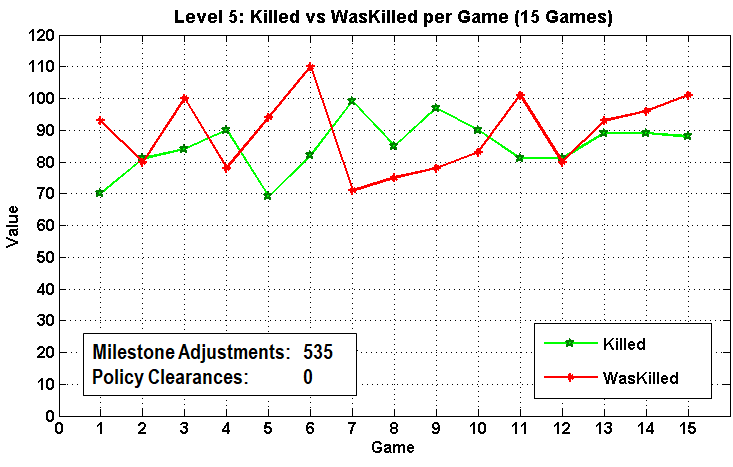}
    \caption{SEC-Bot: Killed vs WasKilled against a Level 5 opponent.}
    \label{secLevel5}
  \end{center}
\end{figure}
This explains the larger variance that is seen in Figure \ref{secLevel5} compared to the other results. It is important to recall that the SEC-Bot was initially trained using this level of opponent and it took almost 50 games before it had enough knowledge and managed to start convincingly defeating the opponent. Even at this point, it was still losing some games.
\subsection{RL-Bot versus Fixed Strategy Opponents}
\begin{figure}[h!]
  \begin{center}  
     \includegraphics[width=3.35in]{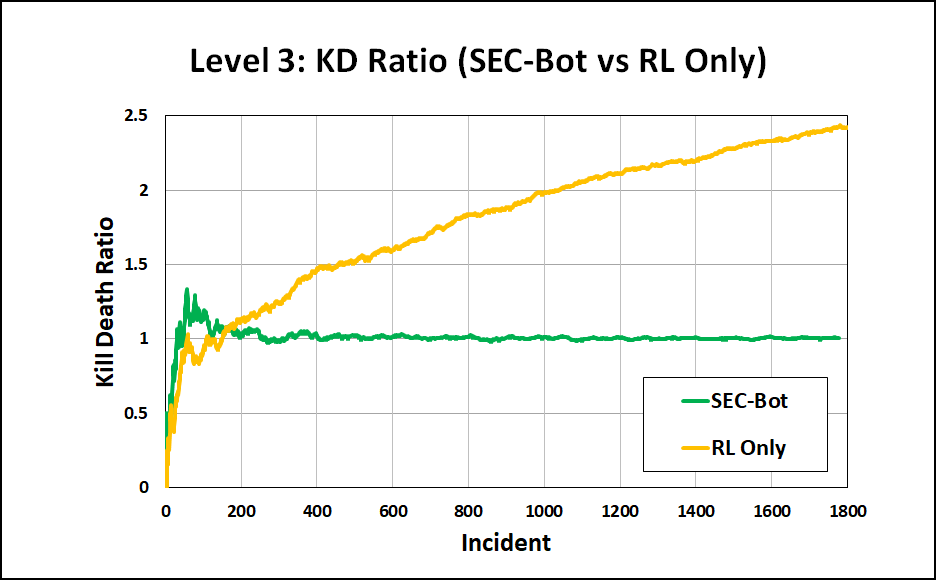}
    \caption{KD Ratio when the SEC mechanism is enabled versus disabled.}
    \label{vs}
  \end{center}
\end{figure}

We also ran all the above experimentation with the SEC mechanism disabled to perform a comparative analysis.  
\begin{table}[h!]
\caption{Comparison of performance when enabling and disabling the SEC mechanism}
\label{secComp}
\begin{center}
\setlength{\extrarowheight}{0.15cm}
\begin{tabular}{|c|c|c||c|c|}
\hline
\vtop{\hbox{\strut \bf{\hspace{0.5mm}}}\hbox{\strut \bf{Level}}} &\cellcolor{orange!25} \vtop{\hbox{\strut \bf{\hspace{0.8mm}RL Only}}\hbox{\strut \bf{KD Ratio}}} & \cellcolor{green!25} \vtop{\hbox{\strut \bf{\hspace{0.8mm}SEC-Bot}}\hbox{\strut \bf{KD Ratio}}}&\cellcolor{orange!25}\vtop{\hbox{\strut \bf{\hspace{2.5mm}RL Only}}\hbox{\strut \bf{Kills$|$Deaths}}}& \cellcolor{green!25} \vtop{\hbox{\strut \bf{\hspace{2.5mm}SEC-Bot}}\hbox{\strut \bf{Kills$|$Deaths}}} \\
\hline
\bf{1}&\cellcolor{orange!25}3.90	&\cellcolor{green!25}1.01		&\cellcolor{orange!25}1955 $|$ 501	&\cellcolor{green!25}988 $|$ 974\\
\hline
\bf{2}	&\cellcolor{orange!25}3.73	&\cellcolor{green!25}1.03		&\cellcolor{orange!25}1915 $|$ 514	&\cellcolor{green!25}976 $|$ 944\\
\hline
\bf{3}	&\cellcolor{orange!25}2.46	&\cellcolor{green!25}1.00		&\cellcolor{orange!25}1561 $|$ 635	&\cellcolor{green!25}1014 $|$ 1011\\
\hline
\bf{4}	&\cellcolor{orange!25}1.33	&\cellcolor{green!25}1.00		&\cellcolor{orange!25}1214 $|$ 915	&\cellcolor{green!25}1004 $|$ 1005\\
\hline
\bf{5}	&\cellcolor{orange!25}0.70	&\cellcolor{green!25}0.99		&\cellcolor{orange!25}1032 $|$ 1476 &\cellcolor{green!25}1302 $|$ 1311\\
\hline
\end{tabular}
\end{center}
\end{table}
The results, recorded over the 15 thirty-minute games for each skill level, are shown in Table \ref{secComp}. The \emph{RL Only} entries represent the reinforcement learning bot with the SEC mechanism disabled. It therefore continues to learn as it gains experience and does not attempt to match the level of the opponent. For each level of opponent, we report the final KD ratio and the number of accumulated kills and deaths over the 15 thirty-minute games for the NPC. From the results, we can see that the performance of the RL Only bot depends on the skill of the opposition whereas the SEC-Bot can adjust its performance to retain a KD ratio of approximately 1.0 in each case.
Figure \ref{vs} shows a comparative visualisation of the KD ratio for the SEC-Bot versus the RL Only bot. 
The ratio is plotted for each game \emph{incident} (an incident occurs when the NPC has killed an opponent or died itself) against the Level 3 opponent. This clearly shows that SEC achieves successful game balancing with an approximate 1:1 KD ratio over time.
\section{Discussion}
The success of our approach relies on the implicit assumption that, during the training phase, the skill-level grows monotonically as the learning time increases. For this reason, the choice of task to be learned is an important one. We have chosen the task of shooting as the NPC becomes naturally more proficient with aiming, through the use of RL, as it encounters more states (movement and orientation of the opponent) over time. The approach is limited by the upper bound of the performance that can be achieved by the NPC and therefore the learning component has to be well defined to avoid potential local minima stagnations during the learning phase. For instance, the SEC-Bot has to adjust to the most knowledgeable policy all of the time in order to compete at the same as a Level 5 fixed-strategy opponent. \\
\indent Another point to note is that our skill balancing technique is only concerned with a single task at the moment, that of weapon proficiency. In this case, tasks such as item collection, enemy avoidance and movement around the map have fixed strategy implementations. We used a small map for both the training and testing phases so that we could focus our evaluation on the shooting performance of the NPC. This is, of course, the key task in an FPS game but there is plenty of scope for learning other secondary tasks in the game and combining them to form the balancing mechanism. \\
\indent There is no guarantee that the underlying RL agent for SEC will exhibit a skillset that will suit all human players, given the different player personas and playing styles that exist. In this research, we decided to focus on the aiming efficiency of the NPC with a single weapon. We ran tests against scripted fixed-strategy opponents so that we could closely monitor the effect of the skill adjustments. We were successful in what we set out to achieve and we believe that it would be straightforward to develop tailor-made behaviours to suit each of the different types of weapon available. Having weapon-specific decision making for multiple weapons could take us a step closer to more generalised behaviour that would suit a wider variety of playing styles.\\
\indent The results for SEC are very promising and show that, using a threshold mechanism and milestones from the learning timeline, we can closely match the level of five different fixed-strategy opponents (with varying degrees of proficiency) using a single instance of the SEC mechanism. We believe that this mechanism could be useful for a wide variety of game genres that produce explicit performance metrics and, in this paper, we have shown how successful it can be in the context of weapon proficiency in an FPS game.
\section{Future Work}
\label{secFutureWork}
The following are some possible refinements that could be applied to the SEC-Bot in order to improve its skill-balancing capability. \\
\indent The criteria for selecting appropriate milestones is an interesting task. Careful performance analysis could aid in the process of milestone creation to determine definitive points of performance improvement which may not follow systematic increments. For instance, we may wish to select a larger number of milestones during the earlier stages of learning and select fewer milestones as the learning begins to plateau. \\
\indent A milestone/player caching system could be introduced for using the SEC against multiple opponents. Each opponent would have an ID and an associated milestone so that the SEC-Bot could switch to an appropriate milestone based on the current opponent. We designed the initial system to balance play with just a single opponent. \\
\indent Finally, the SEC-Bot could be trained against different levels of human players as opposed to fixed-strategy bots. Another approach could be to deploy the SEC-Bot on a server, over the Internet, to train it against both the human and computer-controlled players that it encounters.



%
%
%

\end{document}